\documentclass[conference]{IEEEtran}
\IEEEoverridecommandlockouts
\usepackage{cite}
\usepackage{amsmath,amssymb,amsfonts}
\usepackage{algorithmic}
\usepackage{graphicx}
\usepackage{textcomp}
\usepackage{xcolor}
\def\BibTeX{{\rm B\kern-.05em{\sc i\kern-.025em b}\kern-.08em
    T\kern-.1667em\lower.7ex\hbox{E}\kern-.125emX}}

\usepackage{graphicx}
\usepackage{subcaption}
\usepackage{amsmath}
\usepackage{cite}
\usepackage{multirow}
\usepackage{algorithm}
\usepackage{algorithmic}
\usepackage{enumitem}

\setlist{noitemsep,topsep=2pt,parsep=0pt,partopsep=0pt}
\algsetup{linenosize=\tiny}
    
\begin{document}

\title{Adaptive Thresholding for Visual Place Recognition using Negative Gaussian Mixture Statistics}

\author{
Nick Trinh, Damian Lyons\\
\IEEEauthorblockA{\textit{Dept. of Computer \& Information Science} \\
 \textit{Fordham University}, 
NY USA\\
 \{ntrinhvanminh,dlyons\}@fordham.edu}
}

\maketitle

\begin{abstract}
Visual place recognition (VPR) is an important component technology for camera-based mapping and navigation applications. This is a challenging problem because images of the same place may appear quite different for reasons including seasonal changes, weather illumination, structural changes to the environment, as well as transient pedestrian or vehicle traffic.
Papers focusing on generating image descriptors for VPR report their results using metrics such as \textit{recall@K} and ROC curves. However, for a robot implementation, determining which matches are sufficiently good is often reduced to a manually set threshold. And it is difficult to manually select a threshold that will work for a variety of visual scenarios. 

This paper addresses the problem of automatically selecting a threshold for VPR by looking at the 'negative' Gaussian mixture statistics for a place - image statistics indicating not this place. We show that this approach can be used to select thresholds that work well for a variety of image databases and image descriptors.
\end{abstract}


\section{Introduction}

Visual place recognition (VPR), recognizing when camera imagery indicates that robot is at a location it has seen before, plays an important role in image retrieval, map closure, topological navigation and multi-robot coordination \cite{zaffar2021vpr,Schubert2023VisualPR,lyonsrahouti2023}.
his is often characterized as an image retrieval process: given a database containing images of a set of places under many different weather and lighting conditions and a query image, find the place whose images best match the query image. Part of what makes this a challenging problem is that images of the same place may appear quite different due to seasonal changes, weather illumination, structural changes to the environment, transient pedestrian or vehicle traffic and so forth \cite{Vysotska25}. 

A crucial part of the VPR problem is the processing of images to yield image descriptors, and there has been substantial research on this, e.g., \cite{arandjelovic2016netvlad,Berton2023EigenPlacesTV}.
Often image descriptor work presents results in terms of metrics such as \textit{recall@K} or/and an ROC curve \cite{Schubert2023VisualPR}. Selecting a best match by best score, unfortunately, does not easily allow for the case when the image is not of a known place, since the ranked matches are to known places.
A common approach is to manually select a threshold that distinguishes good matches \cite{Schubert2020UnsupervisedLM}.
However that approach is difficult to implement in a general way.
This paper presents an approach to automatically making that selection in a manner that works well in a variety of cases. The approach is based on looking on considering not what images are representative of a place but rather what are \textit{not representative} of a place. The method is applied to several public image databases and image descriptors to show its general applicability.

The novel research contributions of this paper are:
\begin{itemize}
    \item An approach to automatic threshold selection for VPR on the negative Gaussian mixture statistics of places.
    \item Results showing the method applied to the GardensPoint, SFU and Nordland image databases using Eigenplaces, CosPlace, and AlexNet image descriptors.
    \item Benchmark comparison of the results against previous approaches showing its improvement over baseline and general applicability.
\end{itemize}

The next section will present a review of the state of the art in VPR and thresholds election. Section \ref{sec:method} describes the details of our approach. Section \ref{sec:eval} presents the performance evaluation on public datasets. Section \ref{sec:conc} reports our conclusions and future directions.

\section{Literature Review}


VPR can be cast as database image retrieval \cite{Schubert2023VisualPR}: An image database DB is indexed by a discrete set of places $P$, and when a new query image $q$ (or query image set $Q$) is given, the objective is to find $p_q\in P$ such that $q$ best matches the images for place $p_q$ in DB. A key challenge for practical applications of VPR  is place identification under changing environment conditions and viewpoints.
VPR robotic applications have to process a stream of images \cite{Vysotska25} $q(t)$ and decide using an \textit{image similarity threshold} when places are considered the same  \cite{lyonsrahouti2023}.

The first step in VPR to extract image descriptors for the images in DB and for $q$, and this is typically done using deep learning. CNN-based approach such as AlexNet and more specific VPR CNN approaches such as NetVLAD\cite{arandjelovic2016netvlad} and others \cite{Berton2022RethinkingVG} have shown success in place recognition. Berton et al. \cite{Berton2023EigenPlacesTV} and Ali-Bey et al \cite{mixvpr} train their network by grouping points of view on a place, the latter explicitly grouping and the former calculating the group, hence building in a place-invariance. Zhu et al. \cite{Zhu2023R2FU} show that a Transformer based approach handling both image retrieval and reranking can also be successful for VPR.

With image descriptors in hand, the images in DB can be matched with $q$ yielding a similarity vector $S_q$ of dimension $|DB|$ representing the match score for each image. For image sequences, a similarity matrix $S_Q$ captures inter-sequence similarity. 
Vysotska et al.\cite{Vysotska25} point out that the image similarity threshold, important in both approaches, is frequently picked manually by human expertise. Nonetheless, this is difficult to do in a general fashion, and they focus on adapting the similarity threshold by parameter tuning over a small batch of the sequence for each query.
\cite{Schubert2021} reduces the computation in $S_Q$ by comparing only a small number $q1,q2\in Q$. They calculate a similarity threshold $\theta^{DB}_s$ from the sequence by assuming there are more different places than same places. The outliers in this parameter tuning are the images that are the same places. We take a similar approach of estimating thresholds from different places, but unlike \cite{Vysotska25,Schubert2021} which require per-query or per-sequence parameter tuning, we pre-compute per-place thresholds using Gaussian mixture distributions motivated by the variety of images that must map to a single place.

\section{Automatic Threshold Selection}
\label{sec:method}

The proposed approach assumes an existing method to calculate the image descriptor $\mathbf{d}(i)$ given an image $i$. In our case we use the already available VPR software framework \cite{Schubert2023VisualPR} which includes implementations of several state of the art image descriptors. We assume as input a database DB that includes a set of images indexed by a set of places $P$ where $I_p$ is the set of images for place $p$.

\subsection{Image Query and Per-Place Thresholds}
The first step is to calculate a set of \textit{per-place} thresholds $\Theta = \{ \theta_p : p \in P \}$.
A per-place threshold -- rather than a single similarity threshold for all places -- is based on the observation that the acceptable appearance range for different places varies per place.  
When a query image $q$ is presented, we calculate a cosine similarity for $q$ and each place image $i$ as:
\begin{equation}
    s(q,i) = \frac{\mathbf{d}(q) \cdot \mathbf{d}(i)}{\|\mathbf{d}(q)\| \, \|\mathbf{d}(i)\|}
\end{equation}
where $\mathbf{d}$ is the feature descriptor vectors (or aggregated descriptor representations) for the two images. This metric ranges from $-1$ to $1$, with higher values indicating more similar images. We define the \textit{place similarity vector} as:
    $S = < \max(s(q,i)-\theta_p) : i\in I_p > = < s_p >.$
This can be used to calculate 
\textit{recall@K} or best-match as $\arg\max_p s_p$.

\subsection{Calculate Per-Place Thresholds}

Our observation is that 
the appearance of the same place may have a wide range, and inspecting how similar the same place is to itself may be less useful than inspecting how different the places are to each other. This is similar to the observation by \cite{Schubert2021} in calculating their 
similarity threshold.

The distribution of the threshold for place $p$ is represented by a Gaussian mixture:
\begin{equation}
    M(p) = \sum_{i=1}^n w_i N(\mu_i,\sigma^2_i)
\end{equation}
Each component $i$ of $M(p)$ is calculated by fitting a normal distribution to the cosine similarity scores:
$S_i^- = \{ s(i,j) : j \in I(r), r\in P - \{p\}\}$,
that is, the set of similarities between the $i^{th}$ image of place $p$ and all the other images for places other than $p$; what we have called the negative statistics for $i$. The weight is calculated based on the relative size of each $S^-$ as $w_i = \|S^-_i\| / \sum_{j=1}^n\|S^-_j\|$.

Given $M(p)$, the per-place threshold $\theta_p$ is calculated as:
\begin{equation}\label{eq:weighted}
    \theta_p = \sum_{i=1}^{n} \frac{w_i \tau_i^2}{\sum_{j=1}^{n} \tau_j^2} \mu_i
\end{equation}
where $\tau=1/\sigma$: components with smaller variance contributing more to $\theta_p$.

\section{Implementation}


\begin{algorithm}[t]
\small
\caption{Statistical Threshold Generation \\ GenerateStatisticalThresholds}
\label{alg:threshold_generation}
\begin{algorithmic}[1]
\REQUIRE Dataset $D$ with places $P = \{p_1, p_2, \ldots, p_n\}$, images per place $I = \{i_1, i_2, \ldots, i_m\}$, Number of experimental runs $R$
\ENSURE Place-level thresholds $T_{place}$
\STATE Initialize $image\_scores = \{\}$
\FOR{$run = 1$ to $R$}
    \STATE Generate random train/test split for all places
    \STATE Extract features $F$ for all images using feature extractor
    \FOR{each place $p_i$}
        \FOR{each training image $i_j$ in place $p_i$}
            \STATE Compute mean of bad match scores for image $(p_i, i_j)$
            \STATE Store mean scores for current run
        \ENDFOR
    \ENDFOR
    \STATE Save results for current run
\ENDFOR
\STATE $T_{place} = \text{CalculatePlaceAverages}(image\_scores)$ \\ \{Algorithm 2\}
\RETURN $T_{place}$ 
\end{algorithmic}
\end{algorithm}

\begin{algorithm}[t]
\small
\caption{Calculate Place-Level Averages \\ CalculatePlaceAverages}
\label{alg:calc_place_avg}
\begin{algorithmic}[1]
\REQUIRE $image\_scores$ (mean bad scores per image)
\ENSURE $place\_thresholds$
\STATE Initialize $place\_data = \{\}$ for aggregating results
\FOR{all image $img$ in $image\_scores$}
    \STATE $place = \text{get\_place\_from\_key}(img\_key)$
    \IF{$place$ not in $place\_data$}
        \STATE $place\_data[place] = \{\text{means:[]}\}$
    \ENDIF
    \STATE Append mean score to $place\_data[place][means]$
\ENDFOR
\STATE Initialize $place\_thresholds = \{\}$
\FOR{each $place$ in $place\_data$}
    \STATE $\theta_{place} = \text{mean}(place\_data[place][means])$
    \STATE $place\_thresholds[place] = \theta_{place}$
\ENDFOR
\RETURN $place\_thresholds$
\end{algorithmic}
\end{algorithm}

We built our image matching pipeline by leveraging and modifying the feature extraction component of the VPR\_Tutorial codebase \cite{Schubert2023VisualPR}. This choice should also make our results easy to replicate. The original system offered multiple feature descriptor options; and we evaluated our results on several.
The feature extractors are integrated into the
pipeline such that given a pair of images, the output is two sets of descriptors 
for comparison.

\section{Performance Analysis}
\label{sec:eval}

\subsection{Dataset Preparation}

We developed a "group-and-step" approach for creating mini datasets from sequential traversal data. This method groups consecutive images to form a single place (the "group" parameter) and then skips a specified number of images (the "step" parameter) before defining the next place. For example, with group=3 and step=10, we take 3 consecutive images as one place, skip the next 10 images to ensure spatial separation, then take the next 3 images as the second place, and so on. This ensures that different places are visually distinct and spatially separated by approximately the desired distance 
of $20 m$ for consistency across datasets.

\subsection{Datasets}

We evaluated our approach on three datasets derived from standard VPR benchmarks, each representing different environmental conditions and challenges. 

\textbf{GardensPoint Mini:} Derived from the GardensPoint dataset \cite{Schubert2023VisualPR}, which captures a campus walkway under varying conditions (day left, day right, night right viewpoints). Using the VPR\_Tutorial's built-in subset, we created a mini version with 20 distinct places by grouping 3 consecutive images and skipping 10 images between groups, ensuring approximately 20-meter separation between places. Each place contains 9 images total (3 from each 
condition), providing challenging appearance variations 
for the same physical location.

\textbf{SFU Mini:} Created from the VPR\_Tutorial's subset of the SFU Mountain dataset, which contains multiple traversals of mountain trails. The mini version comprises 192 places with 4 images per place from different traversals under varying weather and lighting conditions. Places are separated by approximately 20 meters using our group-and-step approach. The larger number of places allows us to evaluate scalability while maintaining computational tractability for 
cross-validation.

\textbf{Nordland Mini Variants:} Derived from the Nordland railway dataset, which uniquely provides aligned imagery across four seasons from a train-mounted camera. We created two configurations to explore the effect of grouping strategies:\\
    \noindent\textbf{Nordland\_Mini\_g2s2}: Uses group=2, step=2, resulting in 13,796 places with 8 images each (2 per season), maintaining approximately 20-meter separation.\\
    \noindent\textbf{Nordland\_Mini\_g3s3}: Uses group=3, step=3, resulting in 9,197 places with 12 images each (3 per season), maintaining approximately 20-meter separation.

All datasets were preprocessed to a consistent format with place identifiers following the pattern \texttt{Place\#\#\#\#\_Cond\#\#\_G\#\#}, enabling systematic train-test splitting and per-place threshold calculation.

\subsection{Experimental Setup}

\textbf{Feature Descriptors:} We evaluated our thresholding approach with three different feature extractors to demonstrate generalizability: (1) EigenPlaces \cite{Berton2023EigenPlacesTV}, a state-of-the-art VPR-specific descriptor trained for viewpoint invariance; (2) CosPlace \cite{Berton2022RethinkingVG}, which uses cosine similarity optimization for place recognition; and (3) AlexNet, a general-purpose CNN feature extractor, to establish a baseline with non-specialized features.

\textbf{Threshold Calculation:} We evaluated two threshold methods. The Weighted Average uses Eq. (4) with variance-based weighting ($\tau = 1/\sigma$). The Simple Average computes $\theta_p = \frac{1}{n}\sum \mu_i$, taking an unweighted arithmetic mean of the Gaussian component means, eliminating variance calculations. Since each place has equal image counts in our experiments, $w_i = 1/n$ in both methods.

\begin{table*}[h]
\centering
\caption{Recall performance comparison across datasets and feature descriptors. Best results for each metric shown in bold.}
\label{tab:results}
\begin{tabular}{|l|l|l|c|c|c|c|}
\hline
\textbf{Dataset} & \textbf{Descriptor} & \textbf{Method} & \textbf{Recall@1} & \textbf{Recall@3} & \textbf{Recall@5} & \textbf{Recall@10} \\
\hline
\multirow{9}{*}{GardensPoint Mini (20 places)} & \multirow{3}{*}{EigenPlaces} & Baseline & 61.67 & 96.67 & \textbf{100} & \textbf{100} \\
 & & Simple Avg & \textbf{93.33} & \textbf{98.33} & \textbf{100} & \textbf{100} \\
 & & Weighted Avg & \textbf{93.33} & \textbf{98.33} & \textbf{100} & \textbf{100} \\
\cline{2-7}
 & \multirow{3}{*}{CosPlace} & Baseline & 53.33 & \textbf{96.67} & \textbf{96.67} & \textbf{98.33} \\
 & & Simple Avg & \textbf{93.33} & \textbf{96.67} & \textbf{96.67} & 96.67 \\
 & & Weighted Avg & \textbf{93.33} & \textbf{96.67} & \textbf{96.67} & 96.67 \\
\cline{2-7}
 & \multirow{3}{*}{AlexNet} & Baseline & 43.33 & 78.33 & 88.33 & 93.33 \\
 & & Simple Avg & \textbf{85.00} & \textbf{96.67} & \textbf{98.33} & \textbf{100} \\
 & & Weighted Avg & \textbf{85.00} & \textbf{96.67} & \textbf{98.33} & \textbf{100} \\
\hline
\multirow{9}{*}{SFU Mini (192 places)} & \multirow{3}{*}{EigenPlaces} & Baseline & 76.04 & 89.84 & 95.83 & 97.92 \\
 & & Simple Avg & \textbf{81.25} & \textbf{94.53} & \textbf{97.40} & \textbf{98.96} \\
 & & Weighted Avg & \textbf{81.25} & \textbf{94.53} & \textbf{97.40} & \textbf{98.96} \\
\cline{2-7}
 & \multirow{3}{*}{CosPlace} & Baseline & 62.76 & 81.25 & 87.24 & 92.19 \\
 & & Simple Avg & \textbf{69.27} & \textbf{86.20} & \textbf{90.62} & \textbf{94.79} \\
 & & Weighted Avg & \textbf{69.27} & \textbf{86.20} & \textbf{90.62} & \textbf{94.79} \\
\cline{2-7}
 & \multirow{3}{*}{AlexNet} & Baseline & 57.55 & 70.83 & 74.74 & 82.55 \\
 & & Simple Avg & \textbf{62.50} & \textbf{74.74} & \textbf{79.95} & \textbf{86.72} \\
 & & Weighted Avg & \textbf{62.50} & \textbf{74.74} & \textbf{79.95} & \textbf{86.72} \\
\hline
\multirow{9}{*}{Nordland Mini g3s3 (9,197 places)} & \multirow{3}{*}{EigenPlaces} & Baseline & 69.04 & 83.96 & 87.54 & 91.02 \\
 & & Simple Avg & \textbf{76.72} & \textbf{87.78} & \textbf{90.65} & \textbf{93.51} \\
 & & Weighted Avg & \textbf{76.72} & \textbf{87.78} & \textbf{90.65} & \textbf{93.51} \\
\cline{2-7}
 & \multirow{3}{*}{CosPlace} & Baseline & 68.15 & 85.99 & 89.72 & 93.11 \\
 & & Simple Avg & \textbf{78.21} & \textbf{90.42} & \textbf{92.91} & \textbf{95.43} \\
 & & Weighted Avg & \textbf{78.21} & \textbf{90.42} & \textbf{92.91} & \textbf{95.43} \\
\cline{2-7}
 & \multirow{3}{*}{AlexNet} & Baseline & 39.61 & 49.31 & 53.24 & 58.31 \\
 & & Simple Avg & \textbf{42.85} & \textbf{52.56} & \textbf{56.63} & \textbf{62.26} \\
 & & Weighted Avg & \textbf{42.85} & \textbf{52.56} & \textbf{56.64} & \textbf{62.26} \\
\hline
\multirow{9}{*}{Nordland Mini g2s2 (13,796 places)} & \multirow{3}{*}{EigenPlaces} & Baseline & 69.04 & 83.95 & 87.53 & 91.02 \\
 & & Simple Avg & \textbf{74.42} & \textbf{86.67} & \textbf{89.69} & \textbf{92.77} \\
 & & Weighted Avg & \textbf{74.42} & \textbf{86.67} & \textbf{89.69} & \textbf{92.77} \\
\cline{2-7}
 & \multirow{3}{*}{CosPlace} & Baseline & 68.15 & 85.99 & 89.72 & 93.11 \\
 & & Simple Avg & \textbf{78.21} & \textbf{90.42} & \textbf{92.91} & \textbf{95.43} \\
 & & Weighted Avg & \textbf{78.21} & \textbf{90.42} & \textbf{92.91} & \textbf{95.43} \\
\cline{2-7}
 & \multirow{3}{*}{AlexNet} & Baseline & 39.61 & 49.31 & 53.24 & 58.31 \\
 & & Simple Avg & \textbf{42.85} & \textbf{52.56} & \textbf{56.63} & \textbf{62.26} \\
 & & Weighted Avg & \textbf{42.85} & \textbf{52.56} & \textbf{56.64} & \textbf{62.26} \\
\hline
\end{tabular}
\end{table*}

\subsection{Cross-Validation Procedure}

For each dataset, we performed 50 random train-test splits, where:
One image per place was randomly selected as the test (query) image;
    the remaining images from each place formed the training set;
    per-place thresholds were computed from training images only;
    test images were evaluated against all training images

For each run, we calculated thresholds using the negative statistics approach described in Section III-B. The bad match score distribution for place $p_i$ was computed by comparing training images from $p_i$ against all training images from other places. We then aggregated statistics across runs using both threshold methods to produce final per-place thresholds.

\subsection{Evaluation Metrics}
\label{subsec:setup}
We evaluated performance using Recall@K (K = 1, 3, 5, 10), which measures the percentage of queries where the correct place appears in the top K retrieved results. We compared three retrieval strategies:\\
    \textbf{Baseline:} Standard ranking by similarity score without thresholding, following the Beyond ANN method \cite{Schubert2021} from the VPR\_Tutorial framework\\
    \textbf{Simple Average:} Filter-then-rank using averaged per-place thresholds\\
    \textbf{Weighted Average:} Filter-then-rank using variance-weighted thresholds (as described in Equation 6)

The filter-then-rank approach first eliminates places whose similarity scores fall below their respective thresholds, then ranks the remaining candidates. This two-stage process reduces false positives while maintaining high recall.

\subsection{Results and Discussion}

Table \ref{tab:results} presents the comprehensive evaluation of our adaptive thresholding approach across all datasets and descriptors. 
The \textit{baseline} rows in Table \ref{tab:results} refer to the unchanged performance of \cite{zaffar2021vpr} for each of the image descriptors on our datasets. The simple average and weighted average rows refer to the two methods to calculate the per-place threshold in section \ref{subsec:setup}.


Comparing the baseline rows in Table \ref{tab:results} to the two averages from our approach, we can see that the approach generally improves upon the baseline across datasets and image descriptors. This supports the conclusion that this automatic thresholding approach method can be used to enrich any pipeline. The observant reader will recall that this paper opened by explaining that while Recall@K is without argument a very useful metric, many robot applications need to pick a best match and that is where the manually set threshold comes in. The value of our automatic threshold selection for picking the best match can be seen in the results table under the Recall@1 column. Recall@1 shows  improvement over the baseline in each case.

Comparing the two thresholds calculated in our approach, there is little to choose between them in effectiveness, and the simpler approach is faster, since it does not need variances. 

\section{Conclusions}
\label{sec:conc}

We present an adaptive thresholding approach for visual place recognition that automatically determines per-place acceptance thresholds based on negative Gaussian mixture statistics. Our key contribution is demonstrating that place-specific thresholds, calculated from the distribution of bad match scores,  significantly improves  performance compared to traditional global thresholding or threshold-less  approaches.


This work addresses a critical gap in VPR systems where manual threshold selection has been a persistent challenge, particularly for robotic applications requiring reliable place recognition under varying conditions. Future work will explore dynamic threshold adaptation for online learning scenarios and investigate the method's performance on larger-scale datasets.

\bibliographystyle{IEEEtran}
\bibliography{referencesB}










\end{document}